\documentclass[10pt,twocolumn,letterpaper]{article}

\usepackage{iccv}
\usepackage{times}
\usepackage{epsfig}
\usepackage{graphicx}
\usepackage{amsmath}
\usepackage{amssymb}
\usepackage{diagbox}
\usepackage{multirow}
\usepackage{caption}
\usepackage{subcaption}

\usepackage{booktabs}
\usepackage{multirow}

\usepackage[accsupp]{axessibility}  

\usepackage[pagebackref=true,breaklinks=true,letterpaper=true,colorlinks,bookmarks=false]{hyperref}

\iccvfinalcopy 

\def\iccvPaperID{17} 

\ificcvfinal\fi

\begin{document}

\title{Rethinking Common Assumptions to Mitigate Racial Bias in Face Recognition Datasets}

\author{Matthew Gwilliam$^1$
\quad \quad
Srinidhi Hegde$^1$
\quad \quad
Lade Tinubu$^{1,2}$
\quad \quad
Alex Hanson$^{1}$ \\
$^1$University of Maryland \quad \quad $^2$University of Chicago\\ 
{\tt \small \{mgwillia, srihegde\}@umd.edu \quad lade@uchicago.edu \quad hanson@cs.umd.edu}
}

\maketitle
\ificcvfinal\thispagestyle{empty}\fi

\begin{abstract}

Many existing works have made great strides towards reducing racial bias in face recognition.
However, most of these methods attempt to rectify bias that manifests in models during training instead of directly addressing a major source of the bias, the dataset itself.
Exceptions to this are BUPT-Balancedface/RFW~\cite{wang2019racial} and Fairface~\cite{karkkainen2021fairface}, but these works assume that primarily training on a single race or not racially balancing the dataset are inherently disadvantageous.
We demonstrate that these assumptions are not necessarily valid. 
In our experiments, training on only African faces induced less bias than training on a balanced distribution of faces and distributions skewed to include more African faces produced more equitable models.
We additionally notice that adding more images of existing identities to a dataset in place of adding new identities can lead to accuracy boosts across racial categories. 
Our code is available at \url{https://github.com/j-alex-hanson/rethinking-race-face-datasets}.

\end{abstract}


\section{Introduction}

\begin{figure}[t]
    \includegraphics[width=0.95\linewidth]{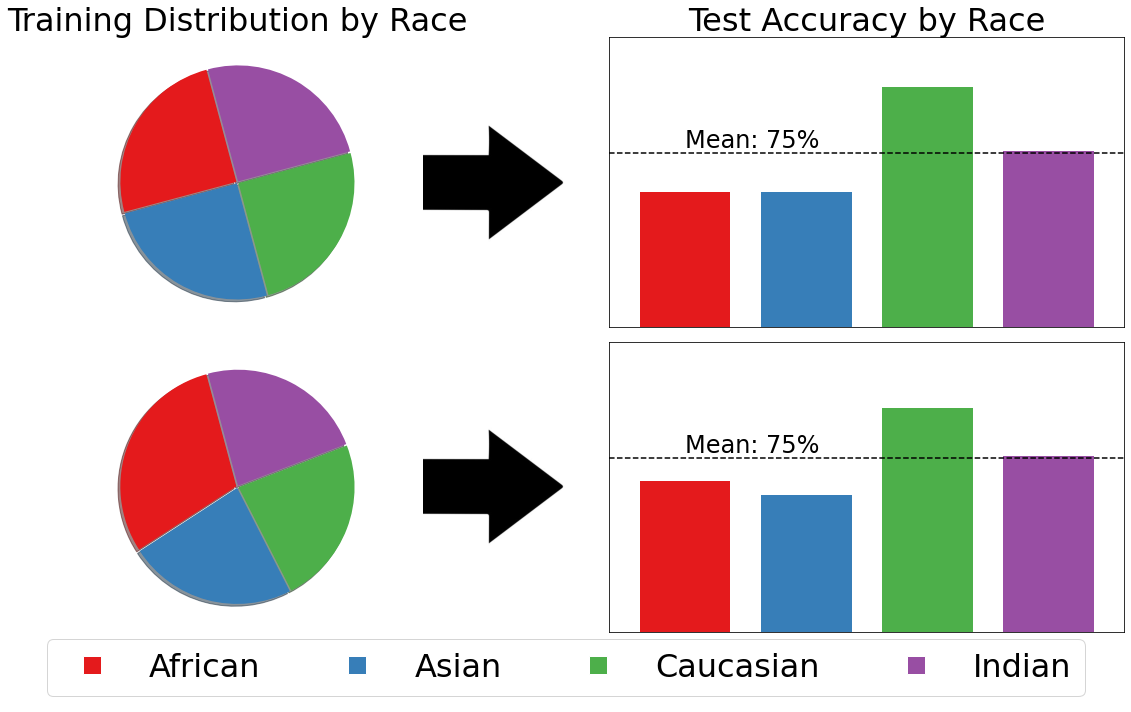}
    \caption{\small 
    \textbf{Balanced vs skewed training dataset.} 
    A training distribution balancing race (all races: $25\%$; top) produces a larger variance on the test accuracies of different races than a skewed distribution (African: $30\%$, other races: $23.3\%$; bottom). 
    The overall accuracy remains at $75\%$, demonstrating that the skewed training distribution is the preferable choice. 
    Data for this figure is from  Section~\ref{sec:race_distributions}.
    See Figure~\ref{fig:extended-teaser} in the Appendix for additional distribution variations.}
    \label{fig:teaser}
\end{figure}

Since before the advent of deep learning, face recognition has been one of the most popular human-centric applications of computer vision.
The introduction of deep learning has only accelerated progress in this area.
The effectiveness of face recognition algorithms makes them a compelling candidate for real-world, industry-level applications.
In fact, face recognition is currently used to unlock phones, validate identities at ATMs, verify drivers licenses, and aid in forensic investigations. 

We consider two key attributes of face recognition applications: performance, which is often measured in the literature in terms of accuracy or a similar metric, and fairness, which is sometimes not measured at all.
The computer vision community has a responsibility to ensure it delivers research, algorithms, and solutions which are not only highly performant, but also very fair.
Therefore, given that existing face recognition applications have high accuracy, we ought to ensure that they are also fair.
One key fairness issue is equitable performance across race. 
Inconsistent performance across race may lead to further disparagement of certain groups, motivating the need for research in this space.

Numerous recent works have attempted to address this concern and show progress towards equability, but this is usually done by modifying the model architecture or data sampling strategy to explicitly induce fairness. 
It is understood that these modifications are necessary because the data used for training these models will inherently cause the models to be biased without some intervention. 
What is it about this data that is causing the models to become biased? 
Two datasets specially created to address racial bias, BUPT-Balancedface (BUPT) / Racial Faces in the Wild (RFW)~\cite{wang2019racial} and Fairface~\cite{karkkainen2021fairface}, posit that the reason bias is introduced is because racial categories are not evenly represented in the data.
This stems from the observation that most commonly used face datasets are primarily comprised of Caucasian individuals. 

But does training on a single race necessarily lead to a biased model? 
And is balancing the dataset across race an optimal choice for mitigating racial bias? 
Figure~\ref{fig:teaser} offers evidence that, contrary to popular assumption, balanced datasets do not always lead to the most balanced results.
The figure shows that for the BUPT dataset -- which groups images into race categories African, Asian, Caucasian, and Indian -- results can be better balanced in terms of accuracy variance, without affecting the mean accuracy, by allocating slightly more images to African that to other race categories.
This example is merely the tip of the iceberg; more thorough analysis follows in the rest of this paper.

There are many more assumptions that are necessary for achieving fairness in face recognition datasets that we do not address in this work.
We do not address the issue of the actual composition and labelling of these datasets. 
BUPT, which we use in this work, has 4 racial categories and every image is placed into one of those 4 categories, sometimes in spite of the represented individual's actual or self-determined race/ethnicity.
Another issue that we do not cover is ensuring that the dataset is capturing a true distribution of faces. 
Face datasets are typically sourced from online images, and images which appear on the internet are in some sense curated -- they were uploaded because the subject or a third party wanted the content to exist online.

Instead, in this work, we analyze assumptions about racial bias in face recognition by analyzing results from the BUPT training dataset and RFW testing dataset. Our contributions are the following:
\begin{itemize}
  \item We show that, when training state-of-the-art models on a single race from BUPT, \textbf{some races generalize across all races differently} than others (Section~\ref{sec:single_race}).
  \item We sample subset training datasets of BUPT from many possible race distributions, and demonstrate that some \textbf{racially skewed datasets mitigate racial bias better than racially balanced datasets}; sometimes by a wide margin (Section~\ref{sec:race_distributions}).
  \item We observe that \textbf{adding more images of existing subjects in face datasets over adding new subjects can lead to greater accuracy boosts} across racial categories (Section~\ref{sec:dataset_size}).
\end{itemize}

For completeness, we define a few terms used throughout our work here. 
We define \textit{face recognition} as the mapping of an image of a face to an identity.
The implementation of face recognition we perform in this work, \textit{pair matching}, requires a trained model (BUPT) to identify if a pair of face images (RFW) belong to the same individual or different individuals; these individuals and their corresponding images are typically unavailable during training.
We define and quantify \textit{racial bias} in face recognition systems as the variance of test race accuracies. For our work, this is the variance in accuracy on the African, Asian, Caucasian, and Indian test splits of RFW.
\label{sec:introduction}

\section{Related Work}

\paragraph{Face Recognition}

Face recognition models have benefited from the inclusion of deep learning techniques~\cite{taigman2014deepface}.
In a typical deep face recognition model, the backbone architecture, discriminative loss function, and a deep feature based face matching method form the three critical components of the system~\cite{wang2018deep}.
A single DNN backbone architecture is the most common choice for the extraction of face features~\cite{parkhi2015deep,schroff2015facenet}.
To exploit these features, face recognition models employ a variety of discriminative loss functions for training face recognition models such as contrastive loss~\cite{wang2017normface}, triplet loss~\cite{schroff2015facenet}, angular/cosine loss~\cite{deng2019arcface,wang2018cosface}, and softmax loss variants~\cite{liu2017sphereface,ranjan2017l2}.
Deep feature based matching methods vary with the requirements of the face recognition application. Face verification can require a more fine-grained approach~\cite{deng2017fine} and face identification necessitates discriminative features \cite{wen2016discriminative}.
In our work, we analyze recent generic face recognition models that are trained with several loss functions and have a single backbone network.

\vspace{-1em}
\paragraph{Bias in Computer Vision for Faces}

In recent years, many studies have confirmed the presence of bias in deep neural networks~\cite{buolamwini2018gender, mehrabi2021survey} which may result in undesired consequences especially for face recognition~\cite{nagpal2019deep, suresh2019framework}.
Several works focus on specifically mitigating bias, either by explicitly changing the models or incorporating data sampling strategies distinctly for this purpose.
Regarding  model modifications, \cite{wang2019racial} propound a model to balance representations of face data from other datasets, in addition to introducing a balance dataset.
\cite{gong2021mitigating} use adaptive convolution kernels and attention mechanisms to mitigate bias in the model.
\cite{serna2020sensitiveloss} incorporate triplet loss to prevent discriminatory effects.
And~\cite{morales2020sensitivenets} propose privacy preserving and learning agnostic representation to mask sensitive information, including race and gender.
Regarding sampling strategies, \cite{wang2020mitigating} uses reinforcement learning dataset sampling to mitigate bias.
\cite{bruveris2020reducing} introduces the use of sampling strategies to mitigate geographic performance differences on photo ID documents. 
\cite{gong2020jointly} uses demographic classifiers and adversarial learning to make face representations more robust. 
\cite{yucer2020exploring} adapts Cycle-GAN to racially balance training per individual in the dataset.
And~\cite{wang2020towards} implements a balanced sampling strategy while training to mitigate bias, though the focus of their work is primarily on gender.

\vspace{-1em}
\paragraph{Datasets}

Face recognition is a problem that has garnered a sustained interest and as such many datasets exist for various subproblems in this space. 
A few popular datasets include: CASIA-WebFace~\cite{yi2014learning}, VGGFace2~\cite{cao2018vggface2}, WebFace260M~\cite{zhu2021webface260m}, MS-Celeb-1M~\cite{guo2016ms}, LFW~\cite{huang2008labeled}, and IJB-C~\cite{maze2018iarpa}. 
Naturally, works also exist that explore potential biases these datasets exhibit~\cite{wen2020adaptive, cao2020domain}. 
Addressing racial bias, datasets such as RFW~\cite{wang2019racial} and FairFace~\cite{karkkainen2021fairface} point out that most of the popular face datasets primarily consist of Caucasian face images and propose a racially balanced face dataset. 
In addition to RFW and FairFace, CASIA-SURF CeFA~\cite{Liu_2021_WACV} is another racially balanced face dataset that addresses racial bias in anti-spoofing.
Data augmentation techniques used to mitigate bias are discussed by~\cite{uchoa2020data,wang2020mitigating,yucer2020exploring}. 
Related to our analysis are~\cite{krishnapriya2020issues}, which finds no evidence that darker skin tone causes higher false match rate on the pair matching problem, and~\cite{albiero2020analysis}, which identifies accuracy differences between genders persist after balancing the dataset. 
\cite{zhang2020class} specifically argues for racially balancing datasets to mitigate bias. 
Our work refutes this claim by demonstrating that some racially skewed training datasets can result in less racially biased models than racially balanced training datasets.
\label{sec:related-work}

\begin{figure}[t]
    \centering
    \includegraphics[width=\linewidth]{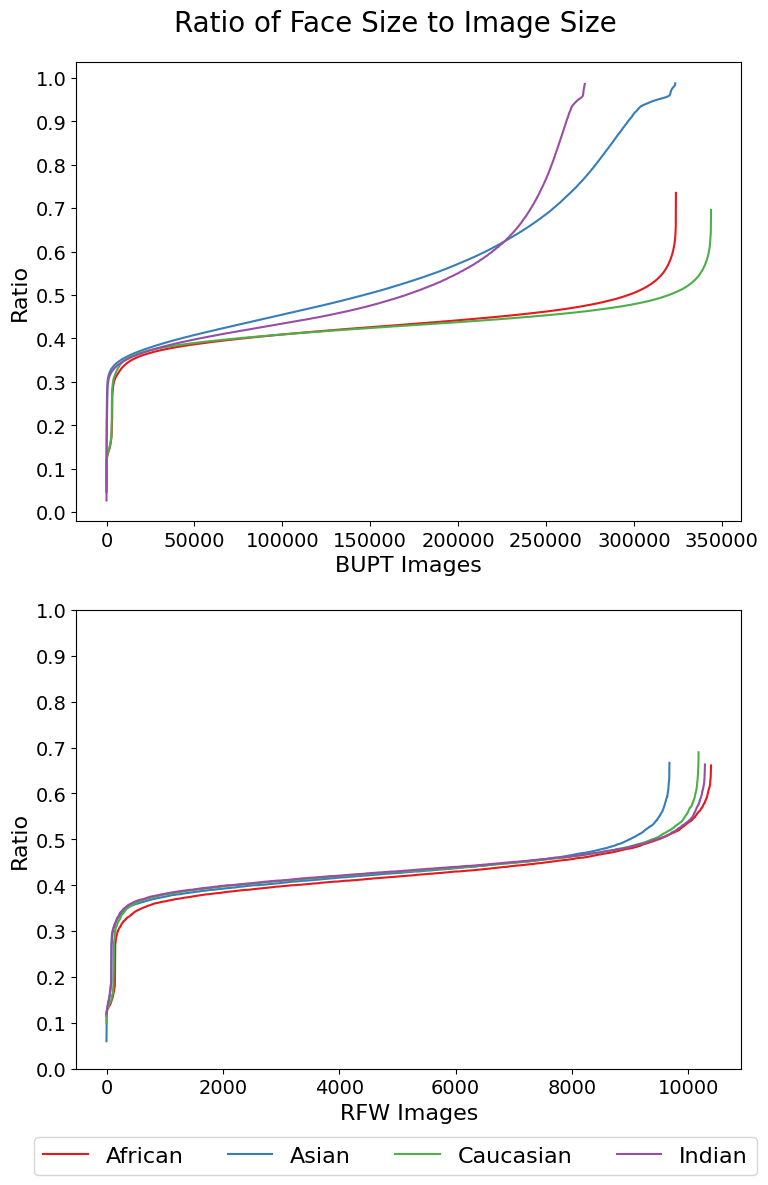}
    \caption{\small \textbf{Face size statistics} for training data (top) and test data (bottom). Both plots show the ratio of the face bounding box area to the area of the containing image, sorted for each race category according to the ratio.}
    \label{fig:face_ratios}
\end{figure}

\begin{figure*}[t]
    \centering
    \includegraphics[width=\linewidth]{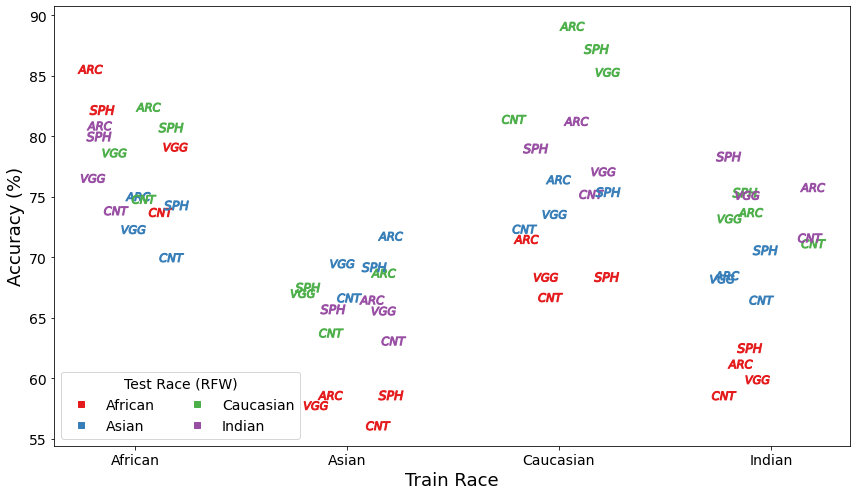}
    \caption{\small \textbf{Results for single race distributions.} Models from ArcFace (ARC), CenterLoss (CNT), SphereFace (SPH), and VGGFace2 (VGG) were trained on each race, and then tested on each race. Results were averaged across 5 trials. Tabular results in Table~\ref{tab:single-race-results}.}
    \label{fig:single_race_results}
\end{figure*}

\section{Experimental Setup}

\subsection{Models}

We run our experiments on four recent face recognition models: VGGFace2, CenterLoss, SphereFace, and ArcFace. 
\textbf{VGGFace2}~\cite{cao2018vggface2} is simply an implementation of a squeeze-and-excitation network (a modified ResNet architecture)~\cite{hu2018squeeze}, trained with a standard cross-entropy loss where the classes are the training dataset identities. 
The outputs of the penultimate layer are used as feature embeddings at test time for pair-matching (see below for more details). 
Because VGGFace2 is a standard network implementation, we view this as a baseline method. 
In addition to cross-entropy loss, \textbf{CenterLoss}~\cite{wen2016discriminative} maintains an additional center vector per identity and imposes a squared L2 distance loss between the feature vectors of the training samples and their corresponding centers. 
The centers are initialized randomly and are incrementally updated by the mean features of each identity as training progresses. 
\textbf{SphereFace}~\cite{liu2017sphereface} modifies the softmax function on the outputs of the network to impose a multiplicative margin loss on the angle between L2 normalized feature embeddings. 
\textbf{ArcFace}~\cite{deng2019arcface} extends SphereFace by including an additive margin loss on this angle between L2 normalized feature embeddings.

Additionally, for fair comparisons between methods we used a ResNet50 as the backbone of each model and trained for 50 epochs. 
All inputs are resized to $128 \times 128$.

\subsection{Datasets}

\begin{table}[ht]
\centering
\setlength{\tabcolsep}{0.25em}
\begin{tabular}{l l l l l l}
\toprule
\multicolumn{2}{c}{Dataset} & \multicolumn{1}{c}{African} & \multicolumn{1}{c}{Asian} & \multicolumn{1}{c}{Caucasian} & \multicolumn{1}{c}{Indian} \\ 
\midrule
\multicolumn{1}{c}{\multirow{2}{*}{\shortstack[c]{BUPT\\ (train)}}} & \# sbjct & \multicolumn{1}{c}{7000} & \multicolumn{1}{c}{7000} & \multicolumn{1}{c}{7000} & \multicolumn{1}{c}{7000} \\ 
{} & \# img & \multicolumn{1}{c}{324376} & \multicolumn{1}{c}{325475} & \multicolumn{1}{c}{326484} & \multicolumn{1}{c}{275095} \\
\cmidrule{1-6}
\multicolumn{1}{c}{RFW (test)} & \# pair & \multicolumn{1}{c}{6000} & \multicolumn{1}{c}{6000} & \multicolumn{1}{c}{6000} & \multicolumn{1}{c}{6000} \\
\bottomrule

\end{tabular}
\caption{\textbf{Data Sources.} Relevant statistics pertaining to the composition of BUPT-Balancedface (BUPT), which is the source for our training datasets, and Racial Faces in the Wild (RFW), which is the source for our testing datasets.}
\label{tab:dataset_stats}
\end{table}

BUPT-Balancedface (BUPT) \cite{wang2019racial}, also known as EqualizedFace in the literature, is the source of our training data.
This dataset, in addition to offering subject/class labels for each image, also groups images into 1 of 4 possible race categories. 
Unlike standard datasets, the distribution of these races in BUPT is balanced, with each race having the same number of subjects, and roughly the same number of images.
For testing, we do a pair-matching task on the pairs in Racial Faces in the Wild (RFW)\cite{wang2019racial}, which is set up such that each race has the same number of corresponding pairs. 
Relevant statistics for both datasets can be found in Table~\ref{tab:dataset_stats}.

We perform additional analysis regarding the contents of these two datasets to produce additional statistics. 
In Figure~\ref{fig:face_ratios}, we show the size distributions of the faces of both datasets, grouped by race. 
We calculated the ratios for each image by dividing the bounding box area of the face, calculated via OpenCV's pre-trained face detector~\cite{opencv_library}, by the area of the containing image.
Notice that while the African and Caucasian size statistics of BUPT are similar to those of all 4 races for RFW, the sizes of the Indian and Asian faces are distributed quite differently.
Specifically, both races are represented by an abundance of larger faces (relative to the images) in the training data.
This could contribute to discrepancies in accuracy and representational power that we observe in Section~\ref{sec:results}.
Moreover, if the balance hypothesis was valid but perturbed by these inconsistent ratios, we would expect an optimally fair distribution to contain an even split between African and Caucasian.
However, our experiments demonstrate this is not the case.

\begin{figure*}
    \centering
    \includegraphics[width=0.98\linewidth]{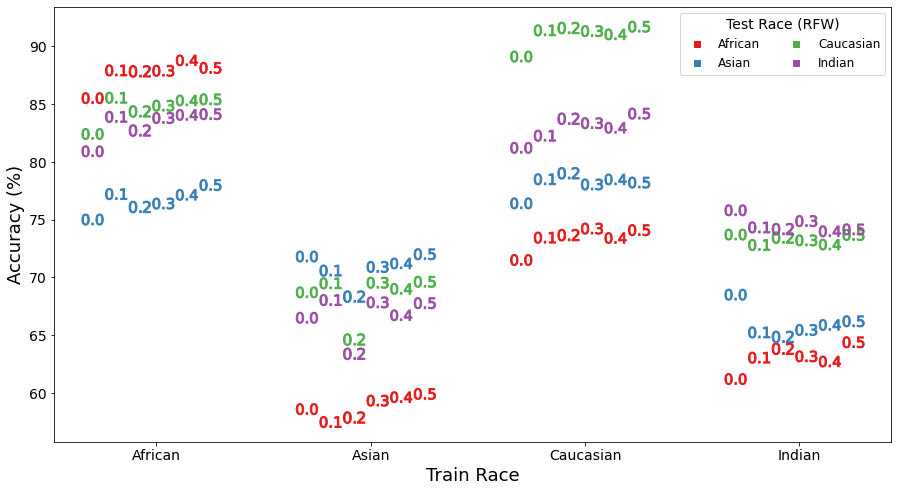}
    \caption{\small \textbf{Effect of noise injection} in training data on model performance. $0.1$, $0.2$, $0.3$, $0.4$, and $0.5$ are the different noise percentage probabilities.}
    \label{fig:noise_analysis}
\end{figure*}

\subsubsection{Single Race Setup}\label{sec:single_race_setup}

For the experiments in Section~\ref{sec:single_race}, we construct our training and testing subsets from the two datasets described above.
Specifically, we take training data from BUPT-Balancedface and testing data from RFW.
Rather than training on the entire BUPT dataset, for the Single Race experiments we train a given model only on a single race.
We repeat this for each race, such that we have different models corresponding to the training data associated with each of the 4 race groups in BUPT.
We evaluate each model on all 4 races in the testing data, separately, such that each model is associated with 4 different test accuracy results, 1 per race.

\subsubsection{Race Distribution Setup}\label{sec:race_dist_setup}

\begin{figure}[ht]
    \centering
    \includegraphics[width=0.85\linewidth]{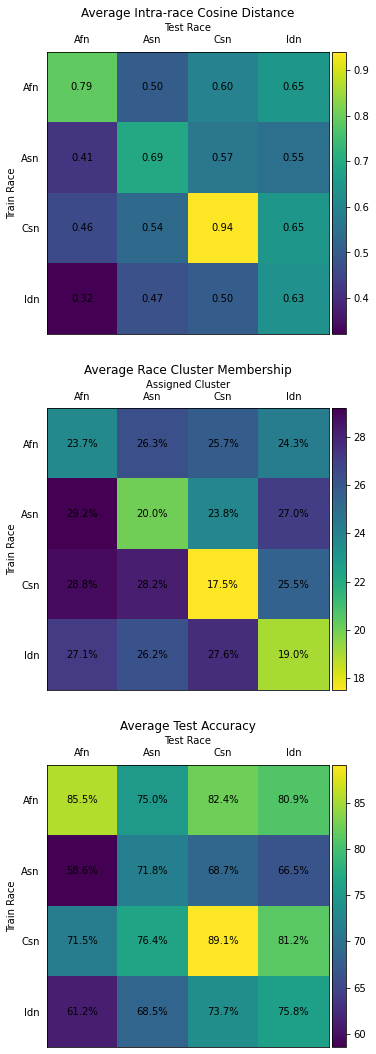}
    \caption{\small \textbf{Cluster vs. accuracy} for the single race experiments using ArcFace.}
    \label{fig:single_race_analysis}
\end{figure}

Prior work on these datasets deals primarily with the uniform (balanced) dataset which consists of equal amounts of persons representing each racial category.
In Section~\ref{sec:race_distributions}, we expand our analysis from Section~\ref{sec:single_race} to a representative collection of non-uniform distributions of data.
Thus, we explore the space of imbalanced datasets.

In order to conduct meaningful exploration, we impose the following constraints:

\begin{itemize}
    \item Each dataset must have the same number of images.
    \item Each dataset must have the same number of subjects.
\end{itemize}

To abide by these constraints, we only consider the 5000 persons corresponding to each race for whom the most data is available. 
Since the least-represented of these persons has 18 corresponding images, we choose to use 18 images to represent each person. 
Furthermore, since we only have 5000 persons available for each race, every dataset considered has exactly 5000 persons, selected from the different races as described below.

We describe each dataset as a tuple consisting of 4 values, $(w_1, x_1, y_1, z_1)$, where $w_1$ gives the portion of the dataset comprised of African persons (as a percent), $x_1$ gives the portion of the dataset that represent Asian persons, and so forth. 
Since the portions are given as percents, the values of each tuple always sum to 100. From the set of all possible tuples, we consider the uniform distribution, $(25, 25, 25, 25)$ and 88 other points. To choose the 88 points, we consider 4 nested 3-simplexes. 
The corners of the outermost simplex are given by permutations of $(100, 0, 0, 0)$. 
The 3 inner simplexes have corners given by permutations of $(60, 13.3, 13.3, 13.3)$, $(40, 20, 20, 20)$, $(30, 23.3, 23.3, 23.3)$ such that we examine more distributions that lie closer to the center (closer to uniform). 
Note that the corners only account for 16 points (4 corners each for 4 simplexes). 
The remaining points are derived by taking 3 equidistant points from each edge on each simplex. 
With 6 edges per simplex, and 4 simplexes, this accounts for the remaining 72 points. 
We believe this method gives reasonable coverage of non-uniform race distributions.

\subsection{Evaluation}

\subsubsection{Test Accuracy}

We compute accuracy for the pair-matching task on RFW. 
This is a binary task, a pair is either a match or not a match. 
Test accuracy is thus the percent of all pairs for a race that are correctly identified as matches or non-matches.

\subsubsection{Cluster Analysis}\label{sec:cluster_analysis_setup}

In order to better understand why different racial image distributions produce different results, we analyze the relationships between features. 
We consider the 4 races as clusters, and assess the image representations by computing metrics that approximate the tightness and spread of the clusters. 
We compute 2 metrics to accomplish this, \textit{Intra-race Cosine Distance} and \textit{Race Cluster Membership}.

\textbf{Intra-race Cosine Distance} 
The first form of ``clustering analysis'' we perform attempts to measure the compactness of the representations corresponding to each race.
We represent images from RFW using the same features used in the pair matching task.
We first compute the mean vector of all the features for the images of a given race. 
Then, we compute the cosine distances between the image features for a race and that race's mean vector. 
We thus report the average of the cosine distances (cluster compactness) from the mean vector (cluster center).

\begin{figure*}[ht]
    \includegraphics[width=\linewidth]{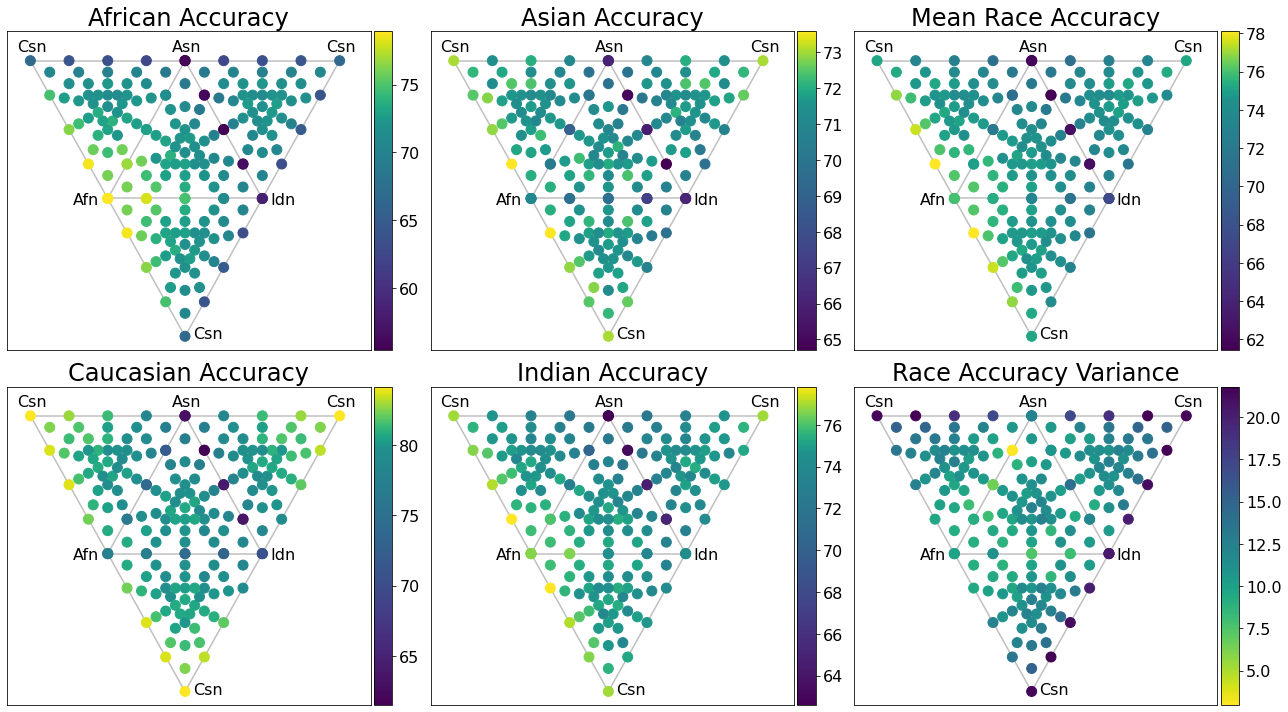}
    \caption{\small \textbf{Race distribution results} of ArcFace models on the 89 distributions identified in Section~\ref{sec:race_dist_setup}. Each plot gives percent accuracy values for a single test race. These plots roughly correspond to flattened versions of the 4 nested 3-simplexes; note that each plot consists of 4 connected equilateral triangles, where each triangle contains 4 parallel simplex faces projected into 2 dimensions, one from each simplex, as well as the center point. The corners of the outer simplex, which correspond to training distributions containing only data from a single race, and are labeled accordingly, i.e., the point labeled ``Afn'' is (100, 0, 0, 0). For readability, some points are therefore represented multiple times, such as the center (25, 25, 25, 25), which appears in the middle of all triangles. Thus, each plot contains 181 points, of which only 89 are unique. In addition to per-race accuracy, overall average accuracy of the races is represented, as well as the accuracy variance between the 4 per-race accuracy results. The variance plot thus represents how ``balanced'' the results are. For reference, the highest mean race accuracy is (75, 0, 25, 0) and the lowest race accuracy variance is (25, 75, 0, 0). Tabular results in Tables \ref{tab:simplex-results-arcface-a} and 
    \ref{tab:simplex-results-arcface-b}.}
    \label{fig:simplex_faces}
\end{figure*}

\textbf{Race Cluster Membership} To complement the cosine distance metric, which is an absolute measure of cluster compactness (treating races as clusters), we perform clustering by taking the 20 nearest neighbors in feature space for each of 5000 randomly selected images from each race of RFW (20000 images in total). 
The neighbors then vote, with the vote weighted according to the inverse of the inner product distance between the image's feature vector and the neighbor's feature vector.
Each images is then assigned the race that receives the maximum votes as the ``cluster'' label.
We repeat this process for each trained model and report the average percent of the RFW images that are assigned to each race.
The resulting cluster membership statistics serve as a relative measure of cluster compactness.
\label{sec:experimental-setup}

\section{Experiments and Analysis}

As Figure~\ref{fig:teaser} establishes, a balanced dataset doesn't actually lend itself to the most balanced results. 
In this section, we explore how training set composition affects results.
In Section~\ref{sec:single_race}, we begin by examining the accuracy results that models trained on a single race achieve when transferred to other races.
Then, in Section~\ref{sec:race_distributions}, we compare models which are trained exclusively on a single race to models which are trained on an even distribution of data to every race, along with 84 distributions in between those extremes, as explained in Section~\ref{sec:experimental-setup}.
Finally, in Section~\ref{sec:dataset_size}, we compare the two different ways to increase the size of a dataset (overall number of images)- gathering more images from existing subjects, and acquiring new subjects.

\subsection{Generalizing from Single Race}\label{sec:single_race}

We train each model (ArcFace, CenterLoss, SphereFace, and VGGFace2) on each race of BUPT (African, Asian, Caucasian, Indian) for 5 trials.
This gives 16 models per trial and 80 models in total.
We test each model on each race of RFW.
Results are averaged across trials.

\subsubsection{Test Accuracy}

Figure~\ref{fig:single_race_results} gives average accuracy results.
Note the major role played by the train race, as it dominates the range of possible accuracy scores.
Also note that, for a given train race, results tend to be grouped not by method, but by train race.
For a given train race and test race, the performance of the various models are relatively similar, with ArcFace tending to have the best results, followed by SphereFace, VGGFace2, and CenterLoss, respectively.
Data is clearly the determining factor in terms of both mean accuracy (across 4 races) as well as fairness (variance in test race accuracies). 

\textbf{Robust Augmentation} 
For understanding the effects of data augmentation on model performance we inject noise in the BUPT train images.
To introduce noise in the image samples, we, first, localize the face using template matching method with Haar features \cite{lienhart2002extended}. 
Then we subdivide the localized face bounding box into a $4\times4$ grid and pick a square patch uniformly at random. 
Finally, we apply Gaussian blur on the selected square patch.
For the Gaussian blur filter, we pick a Gaussian kernel whose size varies randomly between $11$ and $21$ pixels and with a variance of $1.5$.
Throughout this experiment we maintain the original number of training identities, i.e, $7000$.
Figure \ref{fig:noise_analysis} represents the model performance under different noise injection probabilities across the different races which means that each image sample is assigned noise with the specified noise injection probability.
We see that the introduction of noise indeed results in slight boost in accuracy for African and Caucasian races which have stronger representations compared to other races.

\subsubsection{Cluster Analysis}

We attempt to understand how test accuracy is related to the representations learned during training.
As described in Section~\ref{sec:cluster_analysis_setup}, we do this both by considering the races as clusters and obtaining a cosine distance measure for each, as well as clustering via a nearest neighbor voting method.
Figure~\ref{fig:single_race_analysis} shows the results of these methods, in addition to test accuracy, for ArcFace.

We observe reasonable correspondence between the values in the clustering matrices and the accuracy matrix.
Caucasian, from the clustering plots, appears to have the representations which are the most spread out in embedding space, featuring the highest cosine distance as well as the lowest number of assigned image for its own cluster.
African also is fairly spread out according to the cosine distance metric, although it is worth considering that it is somewhat unique in the race cluster membership analysis.
Specifically, African-African is the highest value on the main diagonal, and African training seems to induce more balanced clusters than the other races.
This suggests that while the African embeddings spread out in absolute terms (cosine distance), they also maintain a distinct cluster while also keeping other races clusters distinct.
Overall this seems to empirically suggest that training on races that spread out more in feature space may lead to higher overall accuracy and that more distinct clusters in those spread out embeddings may yield better transfer performance.

\subsection{Analyzing Race Distributions}\label{sec:race_distributions}

It is clear that the distribution of subjects and images among the races affects per-race accuracy.
Furthermore, it is apparent that training on a single race can achieve, in some cases, reasonable performance on other races.
Nevertheless, neither training on a single race nor training on equal amounts of data from all races achieves totally balanced performance across all races.
While, as discussed previously, some researchers attempt to address the remaining imbalances by adopting specific data sampling strategies during training or altering the model embeddings, we study how the dataset itself might be used to balance results.
To more aptly investigate this, we carefully vary the distribution of training data among races, as explained in Section~\ref{sec:race_dist_setup}.
In this way we gain an understanding, not only of the performance of the perfectly balanced dataset, but of a representative portion of the possible distributions of training data among the 4 race supercategories.

Figure~\ref{fig:simplex_faces} shows the results from the selected data distributions with ArcFace.
Notice the importance of training race. For all races, distributions containing a mix of Caucasian and African images tend to be the highest performing.
This matches some of the findings from Figure~\ref{fig:single_race_results}, where Asian accuracy is higher when trained on African or Caucasian images than when trained on Asian images.
Further, the data shown here confirms that the equally balanced dataset, the central point in each plot, is not the best for either performance (mean accuracy) or fairness (variance). 
Instead, as stated before, datasets containing African and Caucasian images tend to give the highest accuracy scores, while various blends that contain African images are the most ``fair'' (lowest variance).
Refer to the Supplementary Material for the race distribution analysis of VGGFace2.

\subsection{Understanding Increasing Dataset Size}\label{sec:dataset_size}

\begin{figure}[h]
    \includegraphics[width=\linewidth]{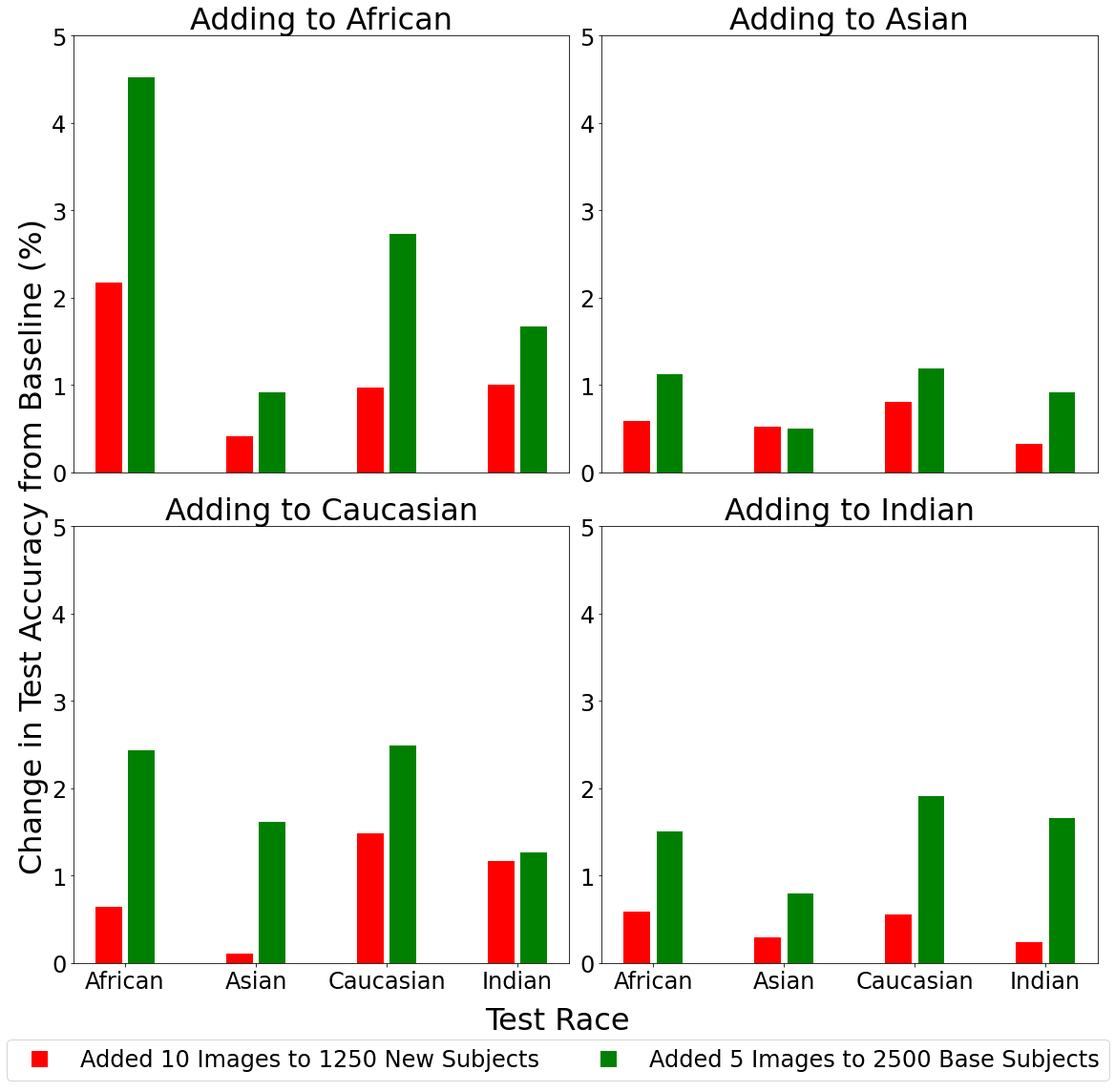}
    \caption{\small \textbf{Increasing dataset size} by adding new subjects vs adding data to existing subjects, for a given race.}
    \label{fig:images_vs_identities_bars}
\end{figure}

This section continues our exploration of the importance of data distribution by investigating how the addition of new data affects results. 
For these experiments, we establish a base dataset, consisting of 10 images per subject for 2500 subjects from each race. 
We examine a special case of data addition where data is added to only one race, and consider two ways this data can be added. 
The first is the introduction of new images for existing subjects; we add 5 images to each of the existing subjects for the selected race. 
The second way is adding new subjects; we introduce 1250 new subjects, and add 10 images to each. 
Both methods for adding data result in the same number of total images, enabling us to compare the two methods for adding to an existing dataset in terms of performance and fairness.

Figure~\ref{fig:images_vs_identities_bars} shows the results of this experiment.
Note that, with the exception of a single train-test pair, accuracy improvements are much higher when images are added to the existing subjects.
Furthermore, adding Caucasian and Indian images in this manner increases accuracy results in a far more balanced fashion than one might expect, with nearly identical improvement for African and Caucasian when adding Caucasian images, and very similar improvement for African, Caucasian, and Indian when adding Indian images.
This reinforces the idea that balancing the distribution of training data can contribute to, but is neither necessary nor sufficient, for balanced results.
\label{sec:results}

\section{Conclusion}

We have shown that common assumptions to mitigate racial bias in datasets do not necessarily hold.
The African BUPT data split produces more equitable models on RFW than a balanced BUPT data split.
Training set distributions skewed to include more African faces also mitigate racial bias better than balanced training sets.
Data augmentations appear to benefit more robust racial categories and adding more images to the base identities of a dataset can boost performance across race.
We have demonstrated some ways to improve mitigation of racial bias on existing datasets, but we hope that illuminating these erroneous assumptions will ultimately assist the face recognition community in building more equitable systems.

\textbf{Acknowledgments:} Alex Hanson was supported by the NDSEG fellowship.\label{sec:conclusion}

{\small
\bibliographystyle{ieee_fullname}
\bibliography{main}
}

\newpage

\appendix

\noindent{\Large \bfseries Appendix}

\section{Additional Single Race Analysis}

In Table~\ref{tab:single-race-results} we provide the results of our single race experiments in tabular form for completeness.

\section{Additional Race Distribution Analysis}

\begin{figure}[ht]
    \includegraphics[width=0.95\linewidth]{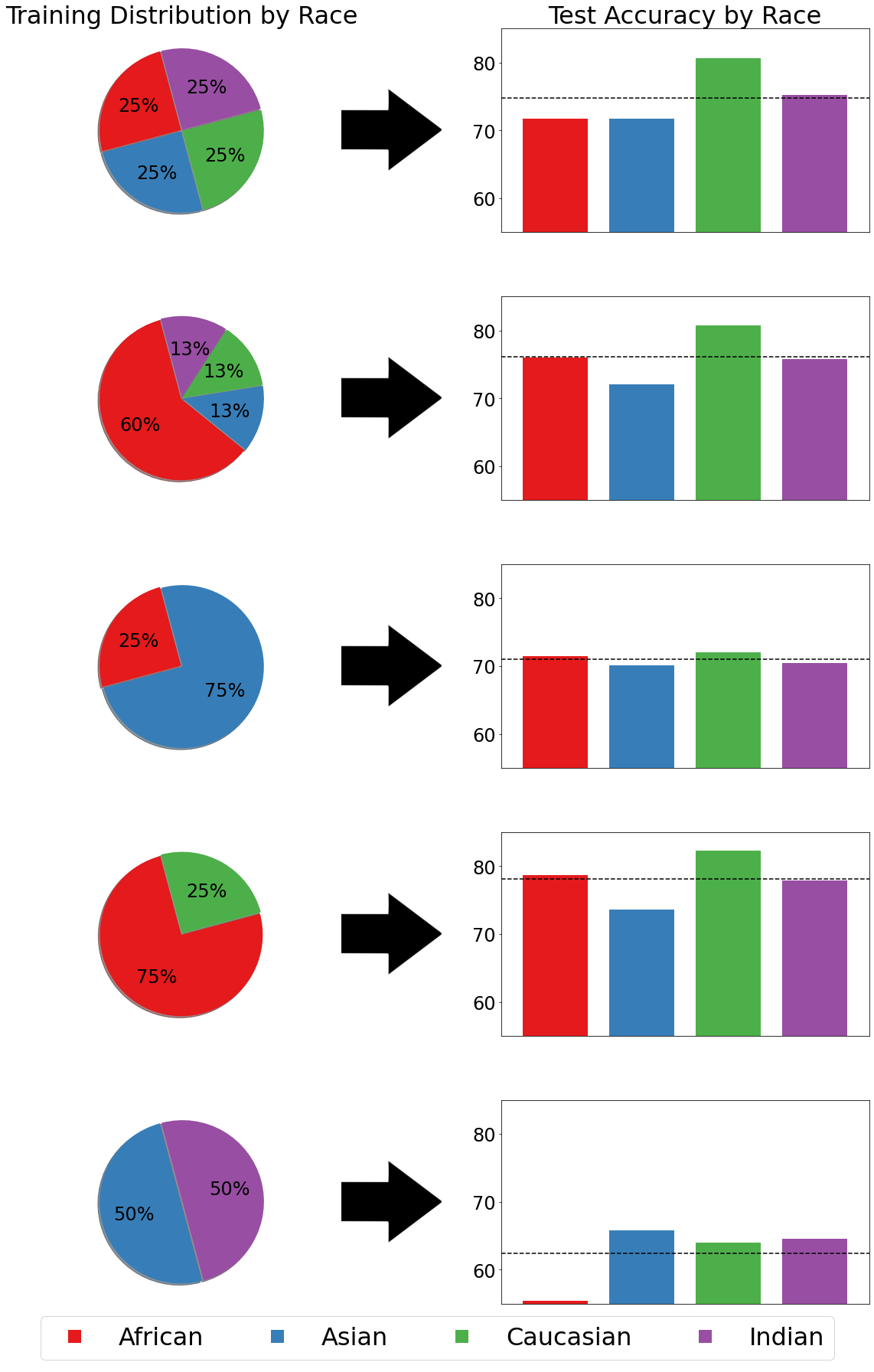}
    \caption{\small \textbf{Balanced vs skewed training datasets.} An extension of Figure~\ref{fig:teaser}, with additional distributions. Data for this figure is described in Section~\ref{sec:race_distributions} provided in Table~\ref{tab:simplex-results-arcface-a} and Table~\ref{tab:simplex-results-arcface-b}}
    \label{fig:extended-teaser}
\end{figure}

In Figure~\ref{fig:extended-teaser} we provide an extended view of Figure~\ref{fig:teaser}, highlighting additional distribution points of interest. This view reinforces that more heavily skewed distributions sometimes do lead to more fairness (less variance) across races.

\begin{figure*}[bp]
    \includegraphics[width=\linewidth]{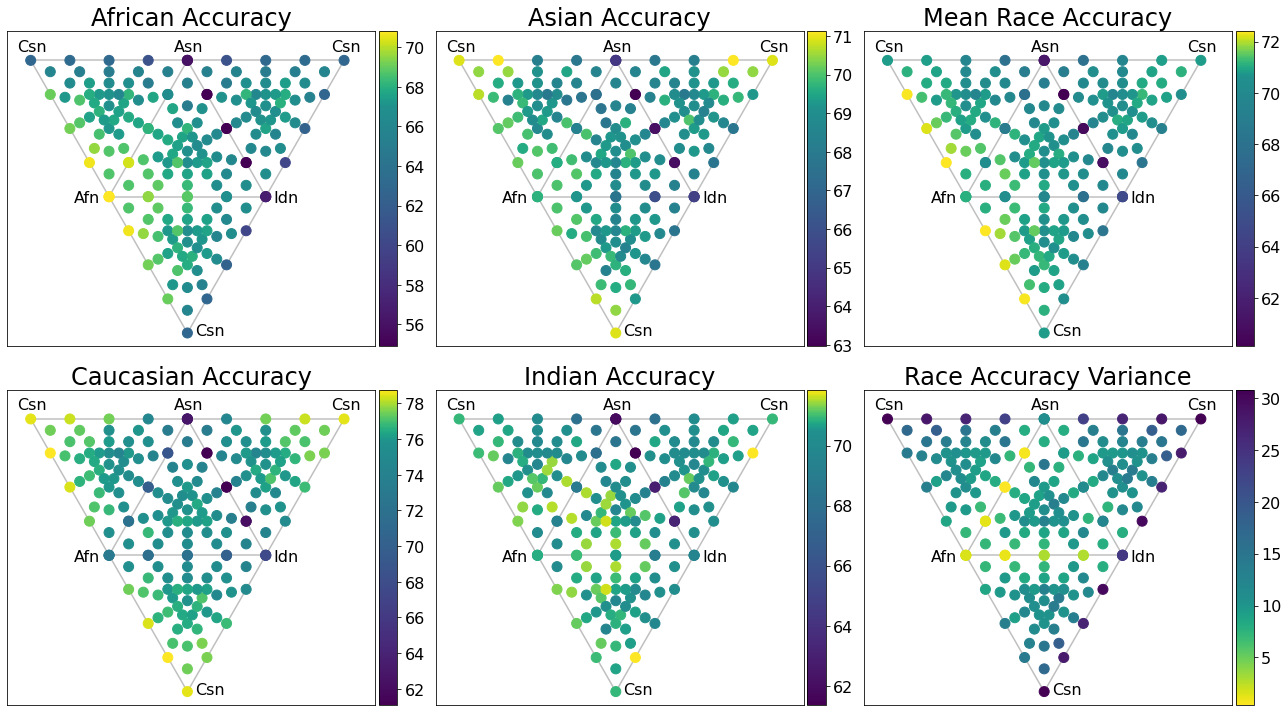}
    \caption{\small
    \textbf{Race distribution results of VGGFace2} on the 89 distributions identified in the main work.
    Each plot gives percent accuracy values for a single test race.
    These plots roughly correspond to flattened versions of the 4 nested 3-simplexes; note that each plot consists of 4 connected equilateral triangles, where each triangle contains 4 parallel simplex faces projected into 2 dimensions, one from each simplex, as well as the center point.
    The corners of the outer simplex, which correspond to training distributions containing only data from a single race, are labeled accordingly, i.e., the point labeled ``Afn'' is (100, 0, 0, 0).
    For readability, some points are therefore represented multiple times, such as the center (25, 25, 25, 25), which appears in the middle of all triangles. 
    Thus, each plot contains 181 points, of which only 89 are unique. 
    In addition to per-race accuracy, overall average accuracy of the races is represented, as well as the accuracy variance between the 4 per-race accuracy results.
    Tabular results in Tables \ref{tab:simplex-results-vggface-a} and 
    \ref{tab:simplex-results-vggface-b}.
    }
    \label{fig:vgg_simplex_faces}
\end{figure*}

In Figure~\ref{fig:vgg_simplex_faces}, we show VGGFace2~\cite{cao2018vggface2} race distribution results are comparable to those shown for ArcFace~\cite{deng2019arcface} in Figure~\ref{fig:simplex_faces}.
The trends they demonstrate are similar in terms of both accuracy and variance (fairness).

In Table~\ref{tab:simplex-results-arcface-a}, Table~\ref{tab:simplex-results-arcface-b}, Table~\ref{tab:simplex-results-vggface-a}, and Table~\ref{tab:simplex-results-vggface-b} we provide the results of our race distribution experiments in tabular form for ArcFace and VGGFace2 respectively (2 tables each) again for completeness.

\begin{table*}[ht]
\centering

\setlength{\tabcolsep}{0.5em}

\begin{tabular}{ l l l l l}
\multicolumn{5}{c}{ArcFace} \\
\toprule

\multirow{2}{*}{Train Race} &   \multicolumn{4}{c}{Test Race} \\ \cmidrule{2-5}
{} & \multicolumn{1}{c}{African} & \multicolumn{1}{c}{Asian} & \multicolumn{1}{c}{Caucasian} & \multicolumn{1}{c}{Indian} \\
\midrule

African		&   85.5 $\pm$ 0.3	& 75.0 $\pm$ 0.2	& 82.4 $\pm$ 0.2	& 80.9 $\pm$ 0.7 \\ 
Asian		& 58.6 $\pm$ 0.5	& 71.8 $\pm$ 0.6	& 68.7 $\pm$ 0.5	& 66.s5 $\pm$ 0.4 \\ 
Caucasian		& 71.5 $\pm$ 0.5	& 76.4 $\pm$ 0.4	& 89.1 $\pm$ 0.2	& 81.2 $\pm$ 0.3 \\ 
Indian		& 61.2 $\pm$ 0.6	& 68.5 $\pm$ 0.5	& 73.7 $\pm$ 0.4	& 75.8 $\pm$ 0.4 \\ \bottomrule
\multicolumn{5}{c}{} \\

\multicolumn{5}{c}{CenterLoss} \\
\toprule

\multirow{2}{*}{Train Race} &   \multicolumn{4}{c}{Test Race} \\ \cmidrule{2-5}
{} & \multicolumn{1}{c}{African} & \multicolumn{1}{c}{Asian} & \multicolumn{1}{c}{Caucasian} & \multicolumn{1}{c}{Indian} \\
\midrule
African		& 73.7 $\pm$ 0.8	& 70.0 $\pm$ 0.6	& 74.8 $\pm$ 0.5	& 73.9 $\pm$ 0.4 \\
Asian		& 56.1 $\pm$ 0.4	& 66.6 $\pm$ 0.6	& 63.7 $\pm$ 0.8	& 63.1 $\pm$ 0.9 \\ 
Caucasian		& 66.7 $\pm$ 0.9	& 72.3 $\pm$ 1.0	& 81.4 $\pm$ 1.1	& 75.2 $\pm$ 0.8 \\
Indian		& 58.6 $\pm$ 1.3	& 66.5 $\pm$ 1.5	& 71.1 $\pm$ 1.2	& 71.6 $\pm$ 0.8 \\
\bottomrule
\multicolumn{5}{c}{} \\

\multicolumn{5}{c}{SphereFace} \\
\toprule

\multirow{2}{*}{Train Race} &   \multicolumn{4}{c}{Test Race} \\ \cmidrule{2-5}
{} & \multicolumn{1}{c}{African} & \multicolumn{1}{c}{Asian} & \multicolumn{1}{c}{Caucasian} & \multicolumn{1}{c}{Indian} \\
\midrule
African		& 82.2 $\pm$ 0.5	& 74.3 $\pm$ 0.4	& 80.7 $\pm$ 0.4	& 80.0 $\pm$ 0.5 \\ 
Asian		& 58.6 $\pm$ 0.5	& 69.2 $\pm$ 0.4	& 67.5 $\pm$ 0.4	& 65.7 $\pm$ 0.5 \\ Caucasian		& 68.4 $\pm$ 0.3	& 75.4 $\pm$ 0.2	& 87.2 $\pm$ 0.3	& 79.0 $\pm$ 0.4 \\ 
Indian		& 62.5 $\pm$ 0.3	& 70.6 $\pm$ 0.3	& 75.3 $\pm$ 0.5	& 78.3 $\pm$ 0.3 \\ 
\bottomrule
\multicolumn{5}{c}{} \\

\multicolumn{5}{c}{VGGFace2} \\
\toprule

\multirow{2}{*}{Train Race} &   \multicolumn{4}{c}{Test Race} \\ \cmidrule{2-5}
{} & \multicolumn{1}{c}{African} & \multicolumn{1}{c}{Asian} & \multicolumn{1}{c}{Caucasian} & \multicolumn{1}{c}{Indian} \\
\midrule
African		& 79.1 $\pm$ 0.4	& 72.3 $\pm$ 0.4	& 78.6 $\pm$ 0.3	& 76.5 $\pm$ 0.5 \\
Asian		& 57.7 $\pm$ 0.6	& 69.5 $\pm$ 0.7	& 67.0 $\pm$ 0.7	& 65.6 $\pm$ 0.5 \\
Caucasian		& 68.4 $\pm$ 0.3	& 73.5 $\pm$ 0.5	& 85.3 $\pm$ 0.4	& 77.1 $\pm$ 0.3 \\
Indian		& 59.9 $\pm$ 0.3	& 68.2 $\pm$ 0.5	& 73.2 $\pm$ 0.2	& 75.1 $\pm$ 0.4 \\ \bottomrule
\multicolumn{5}{c}{} \\

\end{tabular}
\caption{\small \textbf{Results for single race experiments.} We train on BUPT-Balancedface data corresponding to the race in the Train Race row, and test on RFW data corresponding to the race in the Test Race column. These results correspond to those reported in Figure~\ref{fig:single_race_results}; they are provided as a table here for convenience.}
\label{tab:single-race-results}
\end{table*}

\begin{table*}[ht]
\begin{center}
\setlength{\tabcolsep}{0.25em}
\begin{tabular}{ c c c c | c c c c | c c}
\multicolumn{4}{c}{\# Images per Race} & \multicolumn{4}{c}{Per-race Accuracy} & \multicolumn{2}{c}{Overall} \\
African & Asian & Caucasian & Indian & African & Asian & Caucasian & Indian & Accuracy & Variance \\ \hline

1250 & 1250 & 1250 & 1250 & 71.68 & 71.70 & 80.68 & 75.25 & 74.83 & 13.53 \\
1167 & 1167 & 1166 & 1500 & 71.83 & 71.70 & 79.68 & 75.08 & 74.58 & 10.53 \\
1250 & 1167 & 1167 & 1416 & 71.87 & 71.57 & 80.60 & 75.40 & 74.86 & 13.26 \\
1333 & 1167 & 1167 & 1333 & 72.87 & 71.98 & 80.28 & 74.52 & 74.91 & 10.44 \\
1417 & 1167 & 1166 & 1250 & 72.60 & 71.40 & 80.47 & 74.38 & 74.71 & 12.16 \\
1500 & 1167 & 1167 & 1166 & 73.00 & 71.77 & 79.32 & 75.15 & 74.81 & 8.24 \\
1417 & 1250 & 1167 & 1166 & 73.83 & 71.08 & 80.57 & 74.82 & 75.07 & 11.93 \\
1333 & 1333 & 1167 & 1167 & 71.98 & 71.25 & 79.22 & 74.57 & 74.25 & 9.73 \\
1250 & 1417 & 1167 & 1166 & 72.18 & 71.55 & 79.38 & 74.55 & 74.42 & 9.47 \\
1167 & 1250 & 1167 & 1416 & 71.62 & 70.73 & 79.13 & 74.75 & 74.06 & 10.81 \\
1167 & 1333 & 1167 & 1333 & 71.83 & 72.10 & 80.53 & 73.53 & 74.50 & 12.55 \\
1167 & 1417 & 1166 & 1250 & 72.45 & 71.13 & 79.83 & 74.72 & 74.53 & 11.01 \\
1167 & 1500 & 1167 & 1166 & 72.08 & 71.00 & 79.67 & 74.65 & 74.35 & 11.18 \\
1167 & 1417 & 1250 & 1166 & 71.48 & 71.55 & 79.85 & 74.02 & 74.22 & 11.59 \\
1167 & 1333 & 1333 & 1167 & 71.25 & 71.15 & 79.43 & 73.97 & 73.95 & 11.30 \\
1167 & 1250 & 1417 & 1166 & 72.02 & 71.52 & 81.02 & 74.65 & 74.80 & 14.30 \\
1167 & 1167 & 1250 & 1416 & 72.05 & 71.42 & 79.72 & 73.70 & 74.22 & 10.76 \\
1167 & 1167 & 1333 & 1333 & 72.43 & 71.58 & 80.60 & 75.40 & 75.00 & 12.45 \\
1167 & 1167 & 1416 & 1250 & 72.67 & 71.53 & 80.53 & 74.83 & 74.89 & 12.02 \\
1167 & 1167 & 1500 & 1166 & 71.88 & 71.62 & 80.00 & 75.00 & 74.62 & 11.40 \\
1250 & 1167 & 1417 & 1166 & 71.93 & 71.50 & 80.65 & 73.68 & 74.44 & 13.52 \\
1333 & 1167 & 1333 & 1167 & 73.62 & 71.30 & 80.75 & 75.23 & 75.22 & 12.13 \\
1417 & 1167 & 1250 & 1166 & 73.00 & 71.55 & 79.65 & 74.70 & 74.73 & 9.33 \\
\end{tabular}

\begin{tabular}{ c c c c | c c c c | c c}
\multicolumn{4}{c}{\# Images per Race} & \multicolumn{4}{c}{Per-race Accuracy} & \multicolumn{2}{c}{Overall} \\
African & Asian & Caucasian & Indian & African & Asian & Caucasian & Indian & Accuracy & Variance \\ \hline

1250 & 1250 & 1250 & 1250 & 71.68 & 71.70 & 80.68 & 75.25 & 74.83 & 13.53 \\
1000 & 1000 & 1000 & 2000 & 71.08 & 71.78 & 79.65 & 74.33 & 74.21 & 11.32 \\
1250 & 1000 & 1000 & 1750 & 72.02 & 72.40 & 78.62 & 74.38 & 74.35 & 6.86 \\
1500 & 1000 & 1000 & 1500 & 72.68 & 71.22 & 79.90 & 74.37 & 74.54 & 10.81 \\
1750 & 1000 & 1000 & 1250 & 72.92 & 72.15 & 79.53 & 74.82 & 74.85 & 8.24 \\
2000 & 1000 & 1000 & 1000 & 74.53 & 71.85 & 80.33 & 75.57 & 75.57 & 9.40 \\
1750 & 1250 & 1000 & 1000 & 72.17 & 72.28 & 79.30 & 74.12 & 74.47 & 8.39 \\
1500 & 1500 & 1000 & 1000 & 72.98 & 71.65 & 80.33 & 74.32 & 74.82 & 11.02 \\
1250 & 1750 & 1000 & 1000 & 72.27 & 71.20 & 79.72 & 73.98 & 74.29 & 10.80 \\
1000 & 1250 & 1000 & 1750 & 70.83 & 71.15 & 78.83 & 73.47 & 73.57 & 10.26 \\
1000 & 1500 & 1000 & 1500 & 71.80 & 71.00 & 79.42 & 73.68 & 73.98 & 10.82 \\
1000 & 1750 & 1000 & 1250 & 70.48 & 70.70 & 79.20 & 73.28 & 73.42 & 12.36 \\
1000 & 2000 & 1000 & 1000 & 71.18 & 71.62 & 78.80 & 73.57 & 73.79 & 9.17 \\
1000 & 1750 & 1250 & 1000 & 72.02 & 71.47 & 79.42 & 74.08 & 74.25 & 9.86 \\
1000 & 1500 & 1500 & 1000 & 72.28 & 72.42 & 81.07 & 73.58 & 74.84 & 13.19 \\
1000 & 1250 & 1750 & 1000 & 71.80 & 72.47 & 81.73 & 74.72 & 75.18 & 15.49 \\
1000 & 1000 & 1250 & 1750 & 71.88 & 70.85 & 80.00 & 74.47 & 74.30 & 12.57 \\
1000 & 1000 & 1500 & 1500 & 71.62 & 71.05 & 80.67 & 74.12 & 74.36 & 14.58 \\
1000 & 1000 & 1750 & 1250 & 71.80 & 71.78 & 81.40 & 74.78 & 74.94 & 15.40 \\
1000 & 1000 & 2000 & 1000 & 71.55 & 71.28 & 81.37 & 74.78 & 74.75 & 16.51 \\
1250 & 1000 & 1750 & 1000 & 72.35 & 71.78 & 80.52 & 74.85 & 74.88 & 11.94 \\
1500 & 1000 & 1500 & 1000 & 72.60 & 71.68 & 80.63 & 75.82 & 75.18 & 12.26 \\
1750 & 1000 & 1250 & 1000 & 73.82 & 71.48 & 80.35 & 74.58 & 75.06 & 10.64 \\
\end{tabular}
\vspace{-0.5em}
\caption{\small \textbf{Results for ArcFace distribution experiments} for the two innermost simplexes. See Table~\ref{tab:simplex-results-arcface-b} for results for the outermost simplexes. See Figure~\ref{fig:simplex_faces} for these same results in plot form.}
\label{tab:simplex-results-arcface-a}
\end{center}
\end{table*}

\begin{table*}[ht]
\begin{center}
\setlength{\tabcolsep}{0.25em}
\begin{tabular}{ c c c c | c c c c | c c}
\multicolumn{4}{c}{\# Images per Race} & \multicolumn{4}{c}{Per-race Accuracy} & \multicolumn{2}{c}{Overall} \\
African & Asian & Caucasian & Indian & African & Asian & Caucasian & Indian & Accuracy & Variance \\ \hline

1250 & 1250 & 1250 & 1250 & 71.68 & 71.70 & 80.68 & 75.25 & 74.83 & 13.53 \\
667 & 667 & 666 & 3000 & 70.42 & 70.68 & 78.32 & 73.17 & 73.15 & 10.06 \\
1250 & 667 & 666 & 2417 & 72.03 & 71.22 & 79.13 & 75.17 & 74.39 & 9.68 \\
1833 & 667 & 667 & 1833 & 73.52 & 71.60 & 79.68 & 75.27 & 75.02 & 8.94 \\
2417 & 667 & 666 & 1250 & 75.15 & 71.53 & 79.87 & 75.38 & 75.48 & 8.73 \\
3000 & 667 & 667 & 666 & 75.97 & 72.05 & 80.73 & 75.82 & 76.14 & 9.49 \\
2417 & 1250 & 667 & 666 & 75.88 & 71.70 & 79.42 & 75.50 & 75.62 & 7.47 \\
1833 & 1833 & 667 & 667 & 72.72 & 71.63 & 79.57 & 75.25 & 74.79 & 9.32 \\
1250 & 2417 & 667 & 666 & 72.05 & 71.30 & 77.97 & 72.98 & 73.58 & 6.78 \\
667 & 1250 & 666 & 2417 & 70.82 & 71.70 & 78.17 & 73.53 & 73.55 & 8.05 \\
667 & 1833 & 667 & 1833 & 69.80 & 71.10 & 77.90 & 73.17 & 72.99 & 9.47 \\
667 & 2417 & 666 & 1250 & 68.77 & 70.08 & 77.15 & 72.20 & 72.05 & 10.17 \\
667 & 3000 & 667 & 666 & 68.78 & 69.87 & 76.97 & 72.32 & 71.98 & 9.92 \\
667 & 2417 & 1250 & 666 & 70.12 & 71.07 & 79.55 & 73.02 & 73.44 & 13.55 \\
667 & 1833 & 1833 & 667 & 69.68 & 71.32 & 80.33 & 74.08 & 73.85 & 16.47 \\
667 & 1250 & 2417 & 666 & 69.93 & 71.93 & 81.73 & 74.95 & 74.64 & 19.97 \\
667 & 666 & 1250 & 2417 & 70.62 & 70.27 & 79.93 & 73.88 & 73.67 & 15.05 \\
667 & 667 & 1833 & 1833 & 70.17 & 71.67 & 80.93 & 75.00 & 74.44 & 17.11 \\
667 & 666 & 2417 & 1250 & 71.80 & 71.88 & 81.63 & 73.68 & 74.75 & 16.36 \\
667 & 667 & 3000 & 666 & 70.78 & 72.17 & 82.45 & 75.57 & 75.24 & 20.35 \\
1250 & 667 & 2417 & 666 & 72.60 & 72.83 & 81.78 & 76.10 & 75.83 & 13.73 \\
1833 & 667 & 1833 & 667 & 74.52 & 72.47 & 81.58 & 76.12 & 76.17 & 11.44 \\
2417 & 667 & 1250 & 666 & 75.75 & 71.92 & 81.05 & 75.58 & 76.07 & 10.60 \\
\end{tabular}

\begin{tabular}{ c c c c | c c c c | c c}
\multicolumn{4}{c}{\# Images per Race} & \multicolumn{4}{c}{Per-race Accuracy} & \multicolumn{2}{c}{Overall} \\
African & Asian & Caucasian & Indian & African & Asian & Caucasian & Indian & Accuracy & Variance \\ \hline

1250 & 1250 & 1250 & 1250 & 71.68 & 71.70 & 80.68 & 75.25 & 74.83 & 13.53 \\
0 & 0 & 0 & 5000 & 58.38 & 66.17 & 70.57 & 74.32 & 67.36 & 35.17 \\
1250 & 0 & 0 & 3750 & 71.33 & 67.37 & 73.18 & 73.60 & 71.37 & 6.07 \\
2500 & 0 & 0 & 2500 & 74.85 & 69.60 & 74.32 & 74.50 & 73.32 & 4.64 \\
3750 & 0 & 0 & 1250 & 78.03 & 69.87 & 77.10 & 76.58 & 75.40 & 10.46 \\
5000 & 0 & 0 & 0 & 78.92 & 71.05 & 77.28 & 76.65 & 75.97 & 8.77 \\
3750 & 1250 & 0 & 0 & 76.78 & 70.40 & 76.40 & 76.00 & 74.90 & 6.81 \\
2500 & 2500 & 0 & 0 & 72.80 & 68.87 & 74.02 & 72.75 & 72.11 & 3.76 \\
1250 & 3750 & 0 & 0 & 71.52 & 70.15 & 72.05 & 70.43 & 71.04 & 0.60 \\
0 & 1250 & 0 & 3750 & 56.57 & 64.72 & 63.62 & 65.02 & 62.48 & 11.92 \\
0 & 2500 & 0 & 2500 & 55.47 & 65.80 & 63.97 & 64.58 & 62.45 & 16.71 \\
0 & 3750 & 0 & 1250 & 56.12 & 65.03 & 61.52 & 63.20 & 61.47 & 11.09 \\
0 & 5000 & 0 & 0 & 55.73 & 66.03 & 62.98 & 62.60 & 61.84 & 14.19 \\
0 & 3750 & 1250 & 0 & 62.55 & 71.00 & 78.43 & 72.10 & 71.02 & 31.97 \\
0 & 2500 & 2500 & 0 & 63.67 & 72.03 & 81.35 & 73.30 & 72.59 & 39.29 \\
0 & 1250 & 3750 & 0 & 64.23 & 71.50 & 82.80 & 74.70 & 73.31 & 44.41 \\
0 & 0 & 1250 & 3750 & 63.23 & 69.72 & 79.02 & 74.00 & 71.49 & 33.57 \\
0 & 0 & 2500 & 2500 & 64.83 & 70.78 & 81.98 & 74.82 & 73.10 & 38.89 \\
0 & 0 & 3750 & 1250 & 64.13 & 72.58 & 83.27 & 75.25 & 73.81 & 46.66 \\
0 & 0 & 5000 & 0 & 66.48 & 73.05 & 84.15 & 76.92 & 75.15 & 40.91 \\
1250 & 0 & 3750 & 0 & 74.78 & 72.50 & 83.67 & 76.65 & 76.90 & 17.42 \\
2500 & 0 & 2500 & 0 & 76.38 & 72.90 & 83.68 & 77.10 & 77.52 & 15.20 \\
3750 & 0 & 1250 & 0 & 78.68 & 73.58 & 82.30 & 77.83 & 78.10 & 9.61 \\
\end{tabular}
\vspace{-0.5em}
\caption{\small \textbf{Results for ArcFace distribution experiments} for the two outermost simplexes. See Table~\ref{tab:simplex-results-arcface-a} for results for the innermost simplexes. See Figure~\ref{fig:simplex_faces} for these same results in plot form.}
\label{tab:simplex-results-arcface-b}
\end{center}
\end{table*}

\begin{table*}[ht]
\begin{center}
\setlength{\tabcolsep}{0.25em}
\begin{tabular}{ c c c c | c c c c | c c}
\multicolumn{4}{c}{\# Images per Race} & \multicolumn{4}{c}{Per-race Accuracy} & \multicolumn{2}{c}{Overall} \\
African & Asian & Caucasian & Indian & African & Asian & Caucasian & Indian & Accuracy & Variance \\ \hline

1250 & 1250 & 1250 & 1250 & 66.95 & 68.97 & 75.82 & 69.73 & 70.37 & 10.93 \\
1167 & 1167 & 1166 & 1500 & 67.60 & 69.32 & 76.25 & 70.00 & 70.79 & 10.70 \\
1250 & 1167 & 1167 & 1416 & 67.33 & 69.67 & 76.03 & 70.62 & 70.91 & 10.17 \\
1333 & 1167 & 1167 & 1333 & 66.70 & 68.63 & 76.35 & 70.72 & 70.60 & 13.04 \\
1417 & 1167 & 1166 & 1250 & 68.72 & 69.23 & 76.58 & 71.65 & 71.55 & 9.68 \\
1500 & 1167 & 1167 & 1166 & 67.25 & 68.73 & 75.98 & 71.22 & 70.80 & 10.98 \\
1417 & 1250 & 1167 & 1166 & 67.73 & 69.50 & 76.00 & 70.93 & 71.04 & 9.48 \\
1333 & 1333 & 1167 & 1167 & 67.55 & 69.28 & 75.73 & 71.45 & 71.00 & 9.36 \\
1250 & 1417 & 1167 & 1166 & 67.62 & 69.22 & 75.57 & 71.43 & 70.96 & 8.92 \\
1167 & 1250 & 1167 & 1416 & 66.98 & 70.00 & 76.08 & 71.20 & 71.07 & 10.75 \\
1167 & 1333 & 1167 & 1333 & 66.47 & 68.98 & 76.42 & 70.53 & 70.60 & 13.38 \\
1167 & 1417 & 1166 & 1250 & 67.18 & 69.12 & 75.88 & 70.45 & 70.66 & 10.45 \\
1167 & 1500 & 1167 & 1166 & 68.15 & 68.90 & 76.03 & 70.13 & 70.80 & 9.62 \\
1167 & 1417 & 1250 & 1166 & 66.18 & 67.95 & 74.50 & 70.00 & 69.66 & 9.64 \\
1167 & 1333 & 1333 & 1167 & 67.02 & 68.73 & 75.45 & 70.27 & 70.37 & 9.94 \\
1167 & 1250 & 1417 & 1166 & 67.30 & 69.28 & 76.08 & 70.13 & 70.70 & 10.72 \\
1167 & 1167 & 1250 & 1416 & 67.18 & 69.00 & 77.12 & 70.48 & 70.95 & 14.06 \\
1167 & 1167 & 1333 & 1333 & 66.85 & 68.40 & 76.67 & 70.20 & 70.53 & 13.96 \\
1167 & 1167 & 1416 & 1250 & 68.02 & 68.82 & 76.25 & 70.92 & 71.00 & 10.31 \\
1167 & 1167 & 1500 & 1166 & 67.90 & 69.68 & 76.50 & 70.87 & 71.24 & 10.35 \\
1250 & 1167 & 1417 & 1166 & 67.78 & 69.82 & 76.43 & 70.82 & 71.21 & 10.28 \\
1333 & 1167 & 1333 & 1167 & 68.08 & 69.28 & 76.58 & 70.43 & 71.10 & 10.73 \\
1417 & 1167 & 1250 & 1166 & 67.37 & 68.92 & 76.08 & 71.13 & 70.88 & 10.83 \\
\end{tabular}

\begin{tabular}{ c c c c | c c c c | c c}
\multicolumn{4}{c}{\# Images per Race} & \multicolumn{4}{c}{Per-race Accuracy} & \multicolumn{2}{c}{Overall} \\
African & Asian & Caucasian & Indian & African & Asian & Caucasian & Indian & Accuracy & Variance \\ \hline

1250 & 1250 & 1250 & 1250 & 66.95 & 68.97 & 75.82 & 69.73 & 70.37 & 10.93 \\
1000 & 1000 & 1000 & 2000 & 66.32 & 69.25 & 75.75 & 70.33 & 70.41 & 11.66 \\
1250 & 1000 & 1000 & 1750 & 67.52 & 69.25 & 75.47 & 70.68 & 70.73 & 8.74 \\
1500 & 1000 & 1000 & 1500 & 67.58 & 68.45 & 75.20 & 70.98 & 70.55 & 8.76 \\
1750 & 1000 & 1000 & 1250 & 67.75 & 69.52 & 75.52 & 70.70 & 70.87 & 8.30 \\
2000 & 1000 & 1000 & 1000 & 68.73 & 70.03 & 76.82 & 70.78 & 71.59 & 9.64 \\
1750 & 1250 & 1000 & 1000 & 67.33 & 68.88 & 75.65 & 69.92 & 70.45 & 9.87 \\
1500 & 1500 & 1000 & 1000 & 67.80 & 68.52 & 75.30 & 70.33 & 70.49 & 8.57 \\
1250 & 1750 & 1000 & 1000 & 66.67 & 67.80 & 75.60 & 69.75 & 69.95 & 11.84 \\
1000 & 1250 & 1000 & 1750 & 66.52 & 68.98 & 75.32 & 71.12 & 70.48 & 10.44 \\
1000 & 1500 & 1000 & 1500 & 66.33 & 69.05 & 75.78 & 70.08 & 70.31 & 11.85 \\
1000 & 1750 & 1000 & 1250 & 66.88 & 69.65 & 75.10 & 70.30 & 70.48 & 8.75 \\
1000 & 2000 & 1000 & 1000 & 65.00 & 68.50 & 75.60 & 70.75 & 69.96 & 14.79 \\
1000 & 1750 & 1250 & 1000 & 66.82 & 68.52 & 75.68 & 70.07 & 70.27 & 11.09 \\
1000 & 1500 & 1500 & 1000 & 66.45 & 69.15 & 76.08 & 69.97 & 70.41 & 12.41 \\
1000 & 1250 & 1750 & 1000 & 67.38 & 69.25 & 77.18 & 70.15 & 70.99 & 13.77 \\
1000 & 1000 & 1250 & 1750 & 65.88 & 69.27 & 74.83 & 69.72 & 69.93 & 10.23 \\
1000 & 1000 & 1500 & 1500 & 67.33 & 69.20 & 75.98 & 70.68 & 70.80 & 10.36 \\
1000 & 1000 & 1750 & 1250 & 66.67 & 69.63 & 76.72 & 70.63 & 70.91 & 13.36 \\
1000 & 1000 & 2000 & 1000 & 66.50 & 69.58 & 77.07 & 70.93 & 71.02 & 14.77 \\
1250 & 1000 & 1750 & 1000 & 68.67 & 68.32 & 76.55 & 69.65 & 70.80 & 11.28 \\
1500 & 1000 & 1500 & 1000 & 67.57 & 69.87 & 76.90 & 70.57 & 71.23 & 11.97 \\
1750 & 1000 & 1250 & 1000 & 68.55 & 69.93 & 76.87 & 70.43 & 71.45 & 10.27 \\
\end{tabular}
\vspace{-0.5em}
\caption{\small \textbf{Results for VGGFace2 distribution experiments} for the two innermost simplexes. See Table~\ref{tab:simplex-results-vggface-b} for results for the outermost simplexes. See Figure~\ref{fig:vgg_simplex_faces} for these same results in plot form.}
\label{tab:simplex-results-vggface-a}
\end{center}
\end{table*}

\begin{table*}[ht]
\begin{center}
\setlength{\tabcolsep}{0.25em}
\begin{tabular}{ c c c c | c c c c | c c}
\multicolumn{4}{c}{\# Images per Race} & \multicolumn{4}{c}{Per-race Accuracy} & \multicolumn{2}{c}{Overall} \\
African & Asian & Caucasian & Indian & African & Asian & Caucasian & Indian & Accuracy & Variance \\ \hline

1250 & 1250 & 1250 & 1250 & 66.95 & 68.97 & 75.82 & 69.73 & 70.37 & 10.93 \\
667 & 667 & 666 & 3000 & 66.77 & 68.33 & 74.28 & 70.43 & 69.95 & 7.94 \\
1250 & 667 & 666 & 2417 & 66.65 & 68.88 & 75.15 & 71.20 & 70.47 & 9.89 \\
1833 & 667 & 667 & 1833 & 68.32 & 68.92 & 75.22 & 71.52 & 70.99 & 7.40 \\
2417 & 667 & 666 & 1250 & 68.90 & 69.23 & 75.55 & 71.43 & 71.28 & 7.03 \\
3000 & 667 & 667 & 666 & 68.68 & 69.98 & 75.63 & 70.07 & 71.09 & 7.18 \\
2417 & 1250 & 667 & 666 & 68.70 & 69.88 & 75.63 & 71.55 & 71.44 & 6.88 \\
1833 & 1833 & 667 & 667 & 67.68 & 69.97 & 75.45 & 71.52 & 71.15 & 8.01 \\
1250 & 2417 & 667 & 666 & 67.65 & 67.75 & 74.27 & 70.07 & 69.93 & 7.19 \\
667 & 1250 & 666 & 2417 & 66.22 & 68.93 & 73.73 & 70.67 & 69.89 & 7.45 \\
667 & 1833 & 667 & 1833 & 66.28 & 68.62 & 74.20 & 69.30 & 69.60 & 8.30 \\
667 & 2417 & 666 & 1250 & 66.98 & 68.43 & 74.37 & 70.48 & 70.07 & 7.71 \\
667 & 3000 & 667 & 666 & 65.35 & 68.53 & 73.28 & 68.48 & 68.91 & 8.03 \\
667 & 2417 & 1250 & 666 & 65.70 & 69.47 & 75.95 & 70.35 & 70.37 & 13.44 \\
667 & 1833 & 1833 & 667 & 64.92 & 69.12 & 75.75 & 70.27 & 70.01 & 14.94 \\
667 & 1250 & 2417 & 666 & 65.13 & 70.48 & 77.32 & 70.73 & 70.92 & 18.66 \\
667 & 666 & 1250 & 2417 & 66.70 & 68.73 & 75.87 & 70.47 & 70.44 & 11.59 \\
667 & 667 & 1833 & 1833 & 66.43 & 68.60 & 76.33 & 70.42 & 70.45 & 13.54 \\
667 & 666 & 2417 & 1250 & 65.58 & 69.78 & 77.60 & 70.63 & 70.90 & 18.62 \\
667 & 667 & 3000 & 666 & 65.88 & 70.45 & 77.45 & 70.70 & 71.12 & 17.03 \\
1250 & 667 & 2417 & 666 & 67.25 & 69.85 & 77.05 & 71.20 & 71.34 & 12.89 \\
1833 & 667 & 1833 & 667 & 68.77 & 70.15 & 76.67 & 70.75 & 71.58 & 9.13 \\
2417 & 667 & 1250 & 666 & 69.55 & 69.57 & 77.13 & 71.40 & 71.91 & 9.65 \\
\end{tabular}

\begin{tabular}{ c c c c | c c c c | c c}
\multicolumn{4}{c}{\# Images per Race} & \multicolumn{4}{c}{Per-race Accuracy} & \multicolumn{2}{c}{Overall} \\
African & Asian & Caucasian & Indian & African & Asian & Caucasian & Indian & Accuracy & Variance \\ \hline

1250 & 1250 & 1250 & 1250 & 66.95 & 68.97 & 75.82 & 69.73 & 70.37 & 10.93 \\
0 & 0 & 0 & 5000 & 56.75 & 65.33 & 67.72 & 69.63 & 64.86 & 24.24 \\
1250 & 0 & 0 & 3750 & 66.83 & 65.90 & 70.07 & 69.08 & 67.97 & 2.80 \\
2500 & 0 & 0 & 2500 & 68.82 & 67.72 & 72.15 & 70.62 & 69.83 & 2.87 \\
3750 & 0 & 0 & 1250 & 69.53 & 69.20 & 71.73 & 70.98 & 70.36 & 1.08 \\
5000 & 0 & 0 & 0 & 70.80 & 69.68 & 73.13 & 70.82 & 71.11 & 1.58 \\
3750 & 1250 & 0 & 0 & 70.28 & 69.60 & 72.20 & 69.35 & 70.36 & 1.25 \\
2500 & 2500 & 0 & 0 & 68.10 & 68.40 & 69.85 & 68.80 & 68.79 & 0.44 \\
1250 & 3750 & 0 & 0 & 66.60 & 67.45 & 68.97 & 68.00 & 67.75 & 0.74 \\
0 & 1250 & 0 & 3750 & 54.90 & 63.37 & 62.20 & 63.20 & 60.92 & 12.27 \\
0 & 2500 & 0 & 2500 & 54.98 & 63.48 & 61.10 & 62.87 & 60.61 & 11.31 \\
0 & 3750 & 0 & 1250 & 54.92 & 62.97 & 61.18 & 61.37 & 60.11 & 9.47 \\
0 & 5000 & 0 & 0 & 56.02 & 64.97 & 62.75 & 61.88 & 61.40 & 10.94 \\
0 & 3750 & 1250 & 0 & 60.98 & 67.83 & 74.58 & 68.12 & 67.88 & 23.14 \\
0 & 2500 & 2500 & 0 & 63.07 & 68.23 & 77.50 & 70.08 & 69.72 & 26.78 \\
0 & 1250 & 3750 & 0 & 63.25 & 71.12 & 78.28 & 70.65 & 70.82 & 28.28 \\
0 & 0 & 1250 & 3750 & 59.98 & 67.72 & 75.08 & 70.32 & 68.27 & 29.90 \\
0 & 0 & 2500 & 2500 & 62.42 & 67.78 & 76.88 & 69.83 & 69.23 & 26.86 \\
0 & 0 & 3750 & 1250 & 62.87 & 69.23 & 77.57 & 71.85 & 70.38 & 27.89 \\
0 & 0 & 5000 & 0 & 62.83 & 70.90 & 78.52 & 70.93 & 70.80 & 30.76 \\
1250 & 0 & 3750 & 0 & 69.05 & 70.70 & 78.75 & 70.98 & 72.37 & 14.11 \\
2500 & 0 & 2500 & 0 & 69.17 & 70.05 & 78.42 & 71.20 & 72.21 & 13.37 \\
3750 & 0 & 1250 & 0 & 70.65 & 70.18 & 77.45 & 71.32 & 72.40 & 8.66 \\
\end{tabular}
\vspace{-0.5em}
\caption{\small \textbf{Results for VGGFace2 distribution experiments} for the two outermost simplexes. See Table~\ref{tab:simplex-results-vggface-a} for results for the innermost simplexes. See Figure~\ref{fig:vgg_simplex_faces} for these same results in plot form.}
\label{tab:simplex-results-vggface-b}
\end{center}
\end{table*}

\end{document}